\documentclass[letterpaper]{article} 
\usepackage{aaai25}  
\usepackage{times}  
\usepackage{helvet}  
\usepackage{courier}  
\usepackage[hyphens]{url}  
\usepackage{graphicx} 
\usepackage{natbib}  
\usepackage{caption} 
\usepackage{algorithm}
\usepackage{algorithmic}

\usepackage{amsfonts,amssymb}
\usepackage{multirow}
\usepackage{amsmath}
\usepackage{newfloat}
\usepackage{listings}
\DeclareCaptionStyle{ruled}{labelfont=normalfont,labelsep=colon,strut=off} 
\floatstyle{ruled}
\newfloat{listing}{tb}{lst}{}
\floatname{listing}{Listing}
\title{VQLTI: Long-Term Tropical Cyclone Intensity Forecasting with Physical Constraints}
\author{
    Xinyu Wang\textsuperscript{\rm 1},
    Lei Liu\textsuperscript{\rm 1}\thanks{Both are Corresponding Authors.},
    Kang Chen\textsuperscript{\rm 1,\rm 2},
    Tao Han\textsuperscript{\rm 2,\rm 3},
    Bin Li\textsuperscript{\rm 1},
    Lei Bai\textsuperscript{\rm 2}$^*$
}
\affiliations{
    \textsuperscript{\rm 1} University of Science and Technology of China\\
    \textsuperscript{\rm 2} Shanghai Artificial Intelligence Laboratory\\
    \textsuperscript{\rm 3} Hong Kong University of Science and Technology\\
    xinyuwang@mail.ustc.edu.cn, 
    liulei13@ustc.edu.cn, 
    ck6@mail.ustc.edu.cn,
    hantao10200@gmail.com,
    binli@ustc.edu.cn,
    baisanshi@gmail.com
}

\usepackage{bibentry}

\begin{document}

\maketitle

\begin{abstract}

Tropical cyclone (TC) intensity forecasting is crucial for early disaster warning and emergency decision-making. Numerous researchers have explored deep-learning methods to address computational and post-processing issues in operational forecasting. Regrettably, they exhibit subpar long-term forecasting capabilities. We use two strategies to enhance long-term forecasting. (1) By enhancing the matching between TC intensity and spatial information, we can improve long-term forecasting performance. (2) Incorporating physical knowledge and physical constraints can help mitigate the accumulation of forecasting errors. To achieve the above strategies, we propose the VQLTI framework. VQLTI transfers the TC intensity information to a discrete latent space while retaining the spatial information differences, using large-scale spatial meteorological data as conditions. Furthermore, we leverage the forecast from the weather prediction model FengWu to provide additional physical knowledge for VQLTI. Additionally, we calculate the potential intensity (PI) to impose physical constraints on the latent variables. In the global long-term TC intensity forecasting, VQLTI achieves state-of-the-art results for the 24h to 120h, with the MSW (Maximum Sustained Wind) forecast error reduced by \textbf{35.65}$\textbf{\%-}$\textbf{42.51}$\textbf{\%}$ compared to ECMWF-IFS. 

\end{abstract}
\begin{links}
    \link{Code}{https://github.com/1457756434/VQLTI}
\end{links}

%

\section{Introduction}
Tropical cyclones (TCs), commonly known as typhoons or hurricanes, are cyclonic circulations that are stronger in the near-sea surface layer than in the upper layer~\cite{define}. These systems possess tremendous destructive potential, as evidenced by typhoon Mangkhut in 2018, which impacted 3.29 million people and resulted in USD 7.62 billion in direct economic losses~\cite{Mangkhut}. Consequently, accurate long-term forecasting of TC intensity is of critical importance for minimizing disaster-related damages and losses~\cite{flooding,zhang2023forecast}.

\begin{figure}[t]
\centering
\includegraphics[width=0.66\columnwidth]{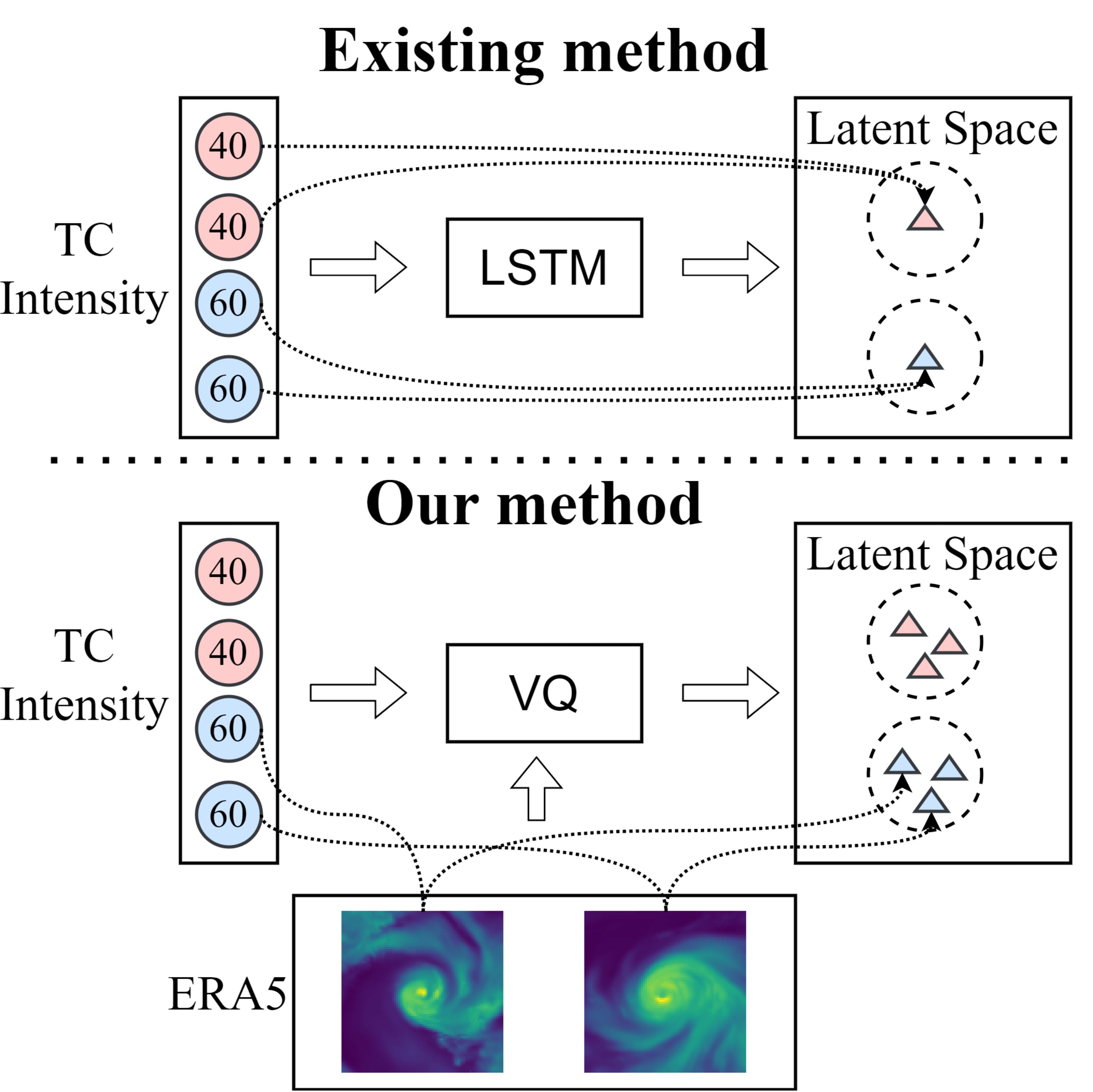}
\caption{Past methods, such as LSTM, would map the same TC intensity to the same position in the latent space, thereby ignoring their spatial information. To preserve this spatial information, we use ERA5 data as a condition to map the TC intensity to a discrete latent space, such that the same TC intensity is distributed with a certain intra-class distance, thereby capturing their spatial differences.}
\label{discrete_latent_space}
\end{figure}

TC intensity forecasting typically refers to forecasting the maximum sustained wind (MSW) and minimum sea level pressure (MSLP) of a TC. In current operational forecasting, these forecasting methods primarily rely on Numerical Weather Prediction models (NWP)~\cite{nwp}, such as the Global Forecast System of the China Meteorological Administration (CMA-GFS)~\cite{cma-gfs}, the Global Forecast System of the National Centers for Environmental Prediction (NCEP-GFS)~\cite{NCEP-GFS}, and the Integrated Forecasting System of the European Centre for Medium-Range Weather Forecasts (ECMWF-IFS)~\cite{ECMWF-IFS}. However, these models usually require the use of high-performance computing resources to solve a large number of partial differential equations, resulting in high computational costs~\cite{cost}. Therefore, the computationally less expensive deep learning (DL) methods have become a research hotspot. Deep learning has made pioneering progress in the field of meteorology, able to understand natural phenomena across temporal and spatial scales, thereby driving the advancement of natural sciences~\cite{pangu,graph_cast,fengwu,fuxi}. In recent years, many researchers have begun to apply deep learning algorithms to the forecasting of TC intensity.

Researchers have conducted some work on deep learning methods for time series forecasting of TC intensity~\cite{RNN-base,transformer-based,mmstn}. They use popular neural network structures like Transformer~\cite{transformer} and LSTM~\cite{LSTM}, based on historical time series information to predict future intensity. However, these simple time series features are still not sufficient to fully represent the current state of the TC, so the forecasting performance is not ideal. More researchers~\cite{TC-Pred,PTCIF,MGTCF,TCIF,MSCAR} have been trying to use multimodal data, such as the renowned fifth generation of ECMWF reanalysis data (ERA5)~\cite{era5}, to effectively extract the spatial features of the TC. This led to decent progress in short-term 24-hour (24h) forecasting. However, these methods generally suffer from significant error accumulation problems, resulting in poorer long-term forecasting performance.

The above methods have two problems. First, there is insufficient matching between TC intensity and spatial information. In the actual dataset, the same TC intensity can exhibit a variety of different spatial patterns. Some past methods~\cite{transformer,mmstn} have neglected the impact of this spatial diversity, which is why relying solely on historical TC intensity performs poorly. Inspired by prior work demonstrating the benefits of data discretization for time series prediction~\cite{vq-ar,vq-tr}, we propose to represent the original intensity in a discrete space. The difference between our method and existing methods is illustrated in Figure~\ref{discrete_latent_space}. By representing the TC intensity information using discrete latent variables, and transforming the original TC intensity forecasting task into forecasting of the corresponding latent variables, this approach enables better matching between the TC intensity information and its spatial patterns.

Secondly, we believe that one important reason for the poor long-term forecasting performance of deep learning models is the insufficient learning of underlying physical knowledge. 
By harnessing the massive global ERA5 dataset, significant advancements have been achieved in machine learning-driven medium-range weather forecasting~\cite{pangu,graph_cast,fengwu,fuxi,fengwu_ghr}. Taking the FengWu model~\cite{fengwu} as an example, it has achieved a skillful 10.75-day forecast horizon for future atmospheric fields. Even more impressively, the TC track forecasting capabilities of FengWu rival or surpass those of the ECMWF-IFS, which is grounded in fluid dynamics equations~\cite{five—wfmodel-error}.
Based on this observation, we believe that the forecast results of FengWu contain certain underlying physical knowledge. If we can effectively utilize this knowledge, it will significantly improve the performance of TC intensity forecasting. Additionally, since the ERA5 dataset itself underestimates TC intensity information~\cite{low1,low2}, the FengWu forecast field also underestimates this information. Supplementing this part of the information could also contribute to performance improvement. Potential intensity (PI) is a theoretical model that characterizes the upper limit of TC intensity under given environmental conditions and constraints~\cite{PI_ori}. We calculate the PI value based on the forecast field, using physical constraints to enhance the forecasting performance.

To address the aforementioned issues, we propose the Vector Quantized Long-term Tropical cyclone Intensity forecasting model (VQLTI). The entire model training involves two stages. In the first stage, with the ERA5 data as a condition, we map the intensity information to a discrete latent space using the Conditional Vector Quantized-Variational AutoEncoder (CVQ-VAE)~\cite{VQVAE} framework. In the second stage, after mapping the historical intensity information to the latent space, similar to LSTM~\cite{LSTM}, we transfer the forecasting task to the latent space. By utilizing the FengWu forecast field to calculate the PI, we apply physical constraints to the forecasting of the latent variables. Then, using the FengWu forecast field as a condition, we decode the forecast latent variables into TC intensity, thereby realizing the forecasting. By adjusting the number of iterations in the latent space, we can achieve forecasts of any time step. The key contributions of this work are:
\begin{itemize} 
\item Proposing to map the TC intensity information to a discrete space, which enables better alignment with the TC spatial information.
\item Incorporating the FengWu forecast field information to achieve high-precision real-time long-term TC intensity forecasting.
\item By calculating the PI, the problem of the forecast field underestimating TC intensity is improved, and the long-term forecasting performance is enhanced.
\item The VQLTI model achieves state-of-the-art (SOTA) performance in global long-term TC intensity forecasting.
\end{itemize}

\begin{figure}[t]
\centering
\includegraphics[width=0.82\columnwidth]{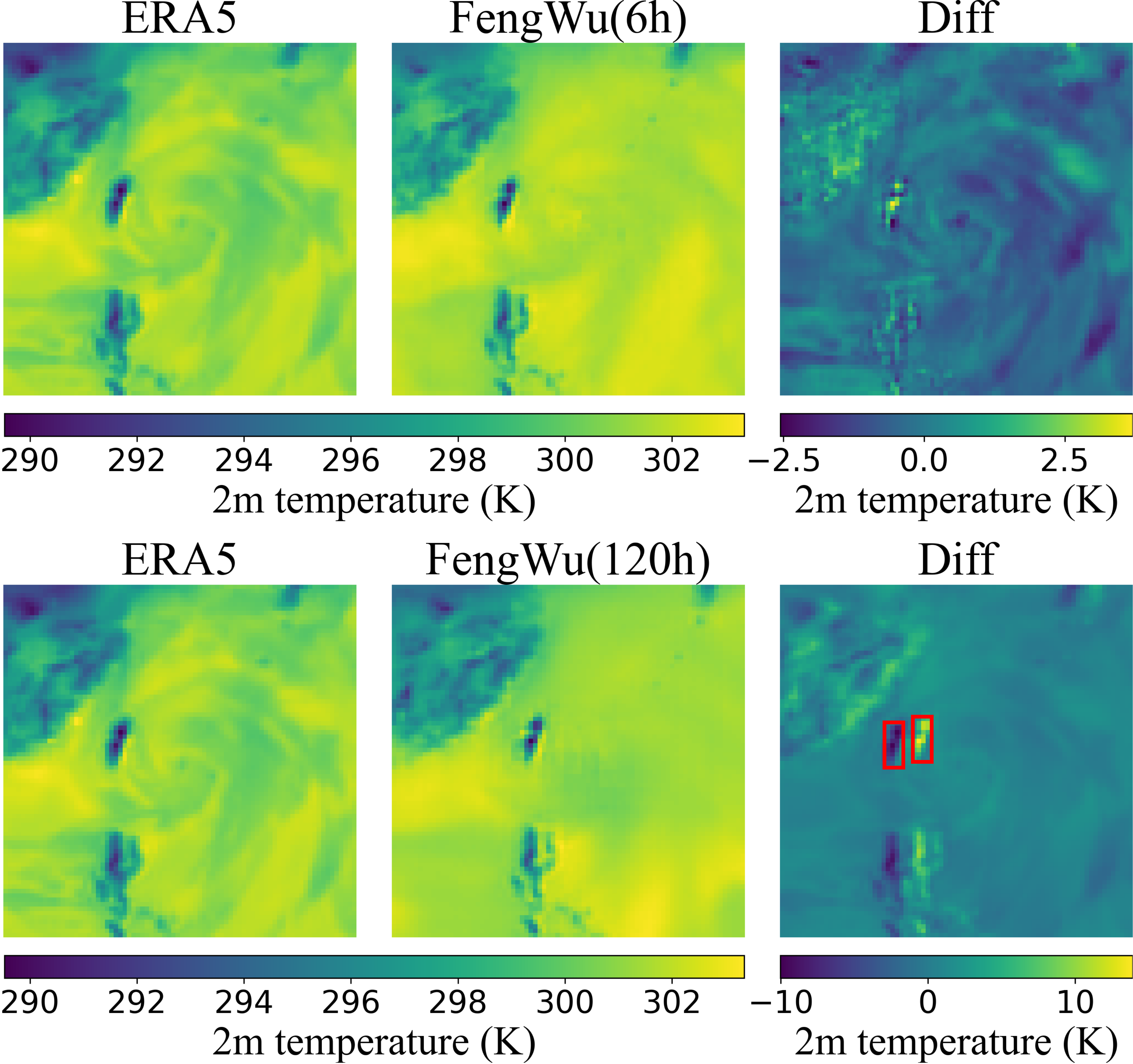}
\caption{For typhoon Hinnamnor at 2022-09-02 18:00, the ERA5 data and FengWu forecast data for 2-meter temperature are visualized, along with their differences. The figure shows FengWu's 6-hour and 120-hour forecasts, with starting times of 2022-09-02 12:00 and 2022-08-28 18:00:00 respectively. As the forecast lead time increases, the tracked TC position from FengWu deviates more, resulting in spatial misalignment between the forecast and actual ERA5 data, as indicated by the red boxes.}
\label{era5_fengwu_diff}
\end{figure}

\section{Preliminaries}
\subsection{Problem Description and Data}
VQLTI takes the past $n$-step TC intensity and ERA5 data, as well as the $m$-step FengWu forecast data as input, to predict the future $m$-step TC intensity. In this study, the ``USA\_WIND'' (knots) and ``USA\_PRES'' (hPa) variables from the IBTrACS dataset~\cite{IBTrACS} are employed as the ground truth for TC intensity, with a temporal resolution of 6 hours, denoted as $I$. 

\begin{figure*}[t]
\centering
\includegraphics[width=0.84\textwidth]{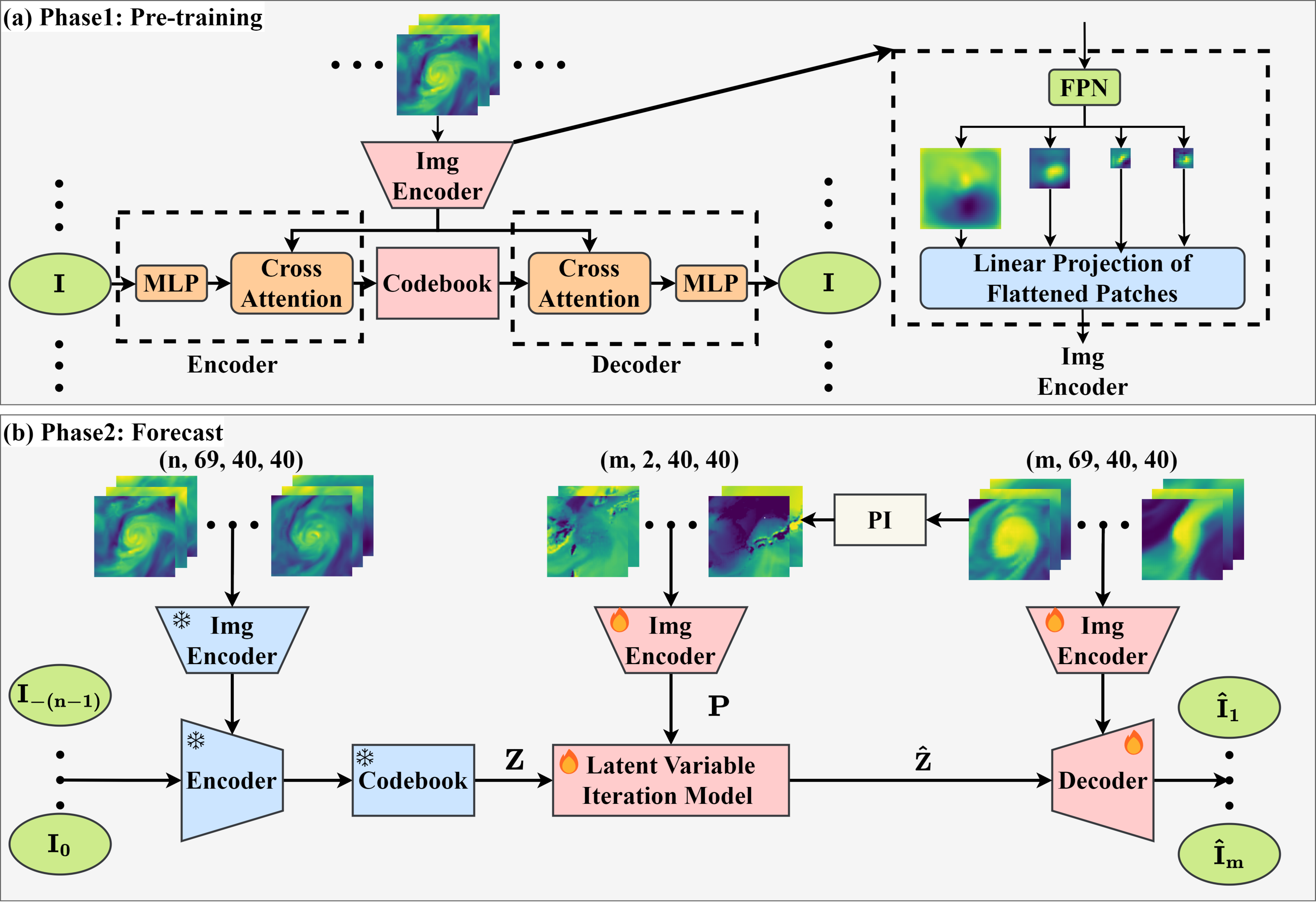} 
\caption{VQLTI Framework. (a) The first stage involves CVQVAE pre-training, which uses ERA5 data as the condition to map TC intensity to a discrete latent space. (b) The second stage is the forecasting stage. Based on the pre-trained model, the blue module is frozen, and the red module is unfrozen. The TC intensity information sequence and the embedding vector in the codebook are matched to obtain the discrete latent variable sequence. When predicting the future latent variables, the Physical Information (PI) is calculated based on the FengWu forecast results to physically constrain the latent variables. Finally, the forecast latent variables are decoded into TC intensity using the FengWu forecast field as the condition.}
\label{VQLTI}
\end{figure*}

The ERA5 data can be interpreted as the fitted data of the actual atmospheric conditions. The same 69 variables as the FengWu model are selected, including 13 levels (50, 100, 150, 200, 250, 300, 400, 500, 600, 700, 850, 925, and 1000 hPa) of 5 variables (geopotential ($z$), specific humidity ($q$), zonal component of wind ($u$), meridional component of wind ($v$), and temperature ($t$)) and 4 surface variables (10-meter u wind component ($u10$), 10-meter v wind component ($v10$), 2-meter temperature ($t2m$), and mean sea level pressure ($msl$))~\cite{fengwu}, and the region with a 10$^{\circ}$ (longitude and latitude)~\cite{TC-Pred} around TC center is cropped. The spatial resolution of the data is 0.25 $^{\circ}$, and the temporal resolution is 6 hours, denoted as $E$.

The FengWu model performs autoregressive forecasting on the historical ERA5 data. Therefore, we utilize the ECMWF's TC tracker~\cite{tracker} to search for the potential future position of the TC on the forecast field and then extract the data within a 10$^{\circ}$ diameter centered on this location, denoted as $F$. Figure ~\ref{era5_fengwu_diff} shows the difference between the FengWu forecast and the ERA5 data, using typhoon Hinnamnor in 2022 as an example.

In summary, this study aims to learn a function $f(\cdot )$ as follows:
\begin{eqnarray}
\begin{aligned}
\hat{I}_{{1:m}} = f(I_{{-(n-1):0}}, E_{{-(n-1):0}}, F_{1:m}, m),
\end{aligned}
\end{eqnarray}
where $\hat{I}_{{1:m}}\in \mathbb{R}^{m\times 2} =\left \{ I_{1}, \cdots, I_{m} \right \}$ denotes the forecast of the future ${m}$ steps of TC intensity, $I_{-(n-1):0}\in \mathbb{R}^{n\times 2}$, $E_{-(n-1):0}\in \mathbb{R}^{n\times 69\times 40 \times 40}$, and $F_{1:m}\in \mathbb{R}^{m\times 69\times 40 \times 40}$.

\subsection{Potential Intensity}
The fundamental energy source for TCs is the release of latent heat from the condensation of water vapor. This process can be described using a Carnot heat engine model to characterize the energy cycle~\cite{PI_ori,PI_name}. Specifically, the surrounding air enters the eyewall, absorbs heat from the ocean, then continuously rises and goes through the energy cycle, before finally flowing out to the outer walls and descending~\cite{PI_100year}. The incorporation of this theoretical model enables the formulation of TC's theoretical maximum intensity, known as the Potential Intensity (PI), which is typically expressed in terms of wind speed as:
\begin{eqnarray}
\begin{aligned}
\left|{V}_{max}\right|^{2} = \frac{C_{k}}{C_{D}} \frac{T_{s}-T_{o}}{T_{o}}\left(k_{0}^{*}-k\right),
\end{aligned}
\end{eqnarray}
where $C_{k}$ represents the dimensionless surface exchange coefficient for enthalpy, $C_{D}$ is momentum surface exchange coefficients, $T_{s}$ denotes the sea surface temperature, $T_{o}$ is the outflow temperature, and $k$ is the enthalpy per unit mass, which is defined as:
\begin{eqnarray}
\begin{aligned}
k \equiv c_{p} T+L_{v} q,
\end{aligned}
\end{eqnarray}
Where $c_{p}$ represents the specific heat capacity at constant pressure, $L_{v}$ denotes the latent heat of vaporization, and $q$ is the specific humidity. In this study, referring to Gilford's calculation process ~\cite{PyPI}, a more concise expression is utilized:
\begin{eqnarray}
\begin{aligned}
V_{max}, p_{min} = PI(t2m, msl, t, q).
\end{aligned}
\end{eqnarray}
The data of $t2m$, $msl$, $t$, and $q$  have been mentioned in ERA5 data. Note that the original equation uses sea-surface temperature, but since FengWu does not perform forecasting, it chooses to use the relatively close t2m as a substitute, and the resulting deviation is considered acceptable.

\section{Methodology}
The VQLTI model consists of two training stages, as Figure~\ref{VQLTI} illustrates. In the first stage, the model employs the CVQ-VAE training framework to learn the mapping from TC intensity information to a discrete latent space. 
The second phase involves learning within the latent space, forecasting future latent variables, and then decoding those into predicted TC intensity.
\subsection{Pre-training}
The first stage of training is inspired by CVAE~\cite{CVAE}, CVAE-GAN~\cite{CVAE-GAN}, and VQ-TR~\cite{vq-tr}, forming an encoder-decoder CVQVAE framework. The Feature pyramid network (FPN)~\cite{FPN} is used to extract multi-scale feature maps of size $\left \{ 40\times40, 20\times20, 10\times10, 10\times10  \right \} $ from the ERA5 data. Similar to the Vision Transformer (ViT)~\cite{ViT}, the feature maps are reshaped into a sequence of flattened 2D patches with $5\times 5$ patch size, which are then concatenated to represent the spatial information of the TC as $E_{p}\in \mathbb{R} ^{88\times d_{p}} $, where $d_{p}$ is the dimension of $E_{p}$. Conditioned on the TC spatial information $E_{p}$, the Encoder maps the TC intensity information to a latent representation. The TC intensity is then passed through a multi-layer perceptron (MLP) and performs cross-attention~\cite{transformer} computation with $E_{p}$. The overall encoding process can be expressed as:
\begin{eqnarray}
    \begin{aligned}
    h & = f_{E}(I\mid E) \\ 
      & = CrossAttn(MLP(I)W_{Q}^{E},E_{p}W_{K}^{E}, E_{p}W_{V}^{E}),
    \end{aligned}
\end{eqnarray}
\begin{eqnarray}
    \begin{aligned}
    CrossAttn(Q,K,V) = softmax(\frac{Q(K )^{T}}{\sqrt{d_{k} } }  )V, 
    \end{aligned}
\end{eqnarray}
where $W_{Q}^{E}$, $W_{K}^{E}$, and $W_{V}^{E}$ are learnable parameter matrices.

Define a learnable discrete latent space $\left \{ e_{1}, \cdots , e_{J} \right \} $ with a size of $J$ and a dimension of $d_{z}$, commonly referred to as the codebook. The encoder output vector $h$ is compared against the discrete vectors in the codebook, and the nearest match is selected as the quantized representation, which can be expressed as:
\begin{eqnarray}
    \label{VQ_fun}
    \begin{aligned}
    Z = e_{q} = VQ(h), \text { where } \quad q = \operatorname{argmin}_{j}\left\|h-e_{j}\right\|_{2}.
    \end{aligned}
\end{eqnarray}
The architectures of the decoder and encoder are analogous. Conditioning on the TC spatial information $E_{p}$, the decoder learns to reconstruct the quantized discrete latent variable $Z$ back to the TC intensity, represented as:
\begin{eqnarray}
    \begin{aligned}
    \hat{I}  & = f_{D}(Z\mid E) \\ 
      & = MLP(CrossAttn(ZW_{Q}^{D},E_{p}W_{K}^{D}, E_{p}W_{V}^{D})).
    \end{aligned}
\end{eqnarray}

Besides the reconstruction loss $L_{recon}$, the overall network also aims to minimize the codebook loss $L_{codebook}$ and the commitment loss $L_{commit}$ introduced by the VQ component. The reconstruction loss, using the mean absolute error (MAE) here, optimizes the encoder and decoder. The codebook loss encourages the encoder's output to align with the codebook vectors. The commitment loss ensures the encoder commits to the embeddings, preventing unbounded output growth and facilitating training convergence. The overall aim is to minimize the following loss function:
\begin{eqnarray}
\begin{aligned}
L&= L_{recon} + L_{codebook} +\beta L_{commit}\\
&= MAE(I, \hat{I} )+\left\|\operatorname{sg}\left[h\right]-e\right\|_{2}^{2}+\beta\left\|h-\operatorname{sg}[e]\right\|_{2}^{2}.
\end{aligned}
\end{eqnarray}
Note that Equation (\ref{VQ_fun}) does not have a gradient, so the stop gradient (sg) operator is used in $L_{codebook}$ and $L_{commit}$. During the forward pass, it is defined as the identity function and has zero partial derivatives. In the actual implementation, the ``detach'' operator blocks the gradient from propagating back. $\beta$ is a hyperparameter.

\subsection{Forecast}
After the pre-training stage, we obtain discrete latent variables that capture the TC intensity information while preserving the associated spatial patterns. If we obtain the discrete latent variables corresponding to the future TC intensity, as well as the future ERA5 data, we can then realize accurate forecasting. Inspired by MSCAR~\cite{MSCAR} and LSTM~\cite{LSTM}, forecasting in the latent space is feasible. The second stage of training will focus on forecasting the latent variables. Based on the pre-trained model, we freeze the encoding and codebook modules, unfreeze the decoding part, and use FengWu's forecasting data to approximate the future ERA5 data, to achieve real-time forecasting. The second stage training process is shown in Figure~\ref{VQLTI}(b). 

The historical TC intensity information is quantified into discrete latent variables window $\left \{ Z_{-(n-1)},\cdots, Z_{0} \right \} $, which are then forecasted through a latent variable iteration model. As TC intensity is influenced by current and historical information, we perform cross-attention calculations on the relevant latent variables, concatenate all the results, and pass them through an MLP layer.

However, this latent space iteration approach suffers from the problem of error accumulation. Additionally, the ERA5 data itself tends to underestimate TC intensity. To address this, we calculate the future PI based on the FengWu forecast data and perform cross-attention between this PI and the MLP output. This provides an upper bound on the possible future TC intensity, which is used to constrain the iterative latent variables physically.

Using a residual connection, we add the last latent variable from the historical window to the one-step latent variable forecast result. We then concatenate this forecast result with the historical window, move the sliding window forward by one step, extract the new latent variables, and repeat the above process to complete the latent space iterative forecasting. The full algorithm is described in Algorithm~\ref{alg:algorithm}.

\begin{algorithm}[tb]
\caption{Latent Variable Iteration Model}
\label{alg:algorithm}
\textbf{Input}: $Z=\left \{ Z_{-(n-1)},\cdots, Z_{0} \right \} , P=\left \{ P_{1},\cdots, P_{m} \right \} $\\
\textbf{Output}: $\hat{Z}=\left \{ \hat{Z} _{1},\cdots, \hat{Z} _{m} \right \} $
\begin{algorithmic}[1] 
\STATE \textbf{Initialize window}: $W=\left \{ Z_{-(n-i)},\cdots, Z_{i-1} \right \}$
\FOR{$i=0;i < m;i++$}
    \STATE $score=CrossAttn(W[0], W[0], W[0])$
    \FOR{$j=1;j < n;j++$}
        \FOR{$k=0;k < (j+1);k++$}
            \STATE$score_j=CrossAttn(W[k], W[j], W[j])$ 
            
            \STATE$score \gets $ Concat $score$ and $score_j$

        \ENDFOR
    \ENDFOR
    \STATE$score_{i}^{future}=MLP(score)$
    \STATE$\hat{Z}[i]=CrossAttn(score_{i}^{future}, P[i], P[i])$
    \STATE$W \gets $ Concat $W$ and $\hat{Z}[i]$
    \STATE$W=W[1:] \gets $Slide one step
\ENDFOR
\STATE \textbf{return} $\left \{ \hat{Z} _{1},\cdots, \hat{Z} _{m} \right \} $
\end{algorithmic}
\end{algorithm}

For the forecasting task, it's important to note that using future ERA5 data is impossible for real-time forecasting. 
Therefore, the final approach is to use the FengWu forecast as the conditioning input and decode all the forecast latent variables $\left \{ \hat{Z} _{1},\cdots, \hat{Z} _{m} \right \} $ back to the TC intensity.

\section{Experiments}
We test the 5-day long-term forecast of global TC intensity and achieve SOTA performance. To assess its practical applicability, we further conduct real-time forecasting experiments, taking into account the actual operating conditions. Finally, we carry out ablation studies to validate the rationality of our model design.
\subsection{Experimental Settings}

\subsubsection{Data Description.}
The experiment uses global TC-related data from the year 1980 to 2022. As mentioned earlier, the IBTrACS dataset~\cite{IBTrACS} is used as the ground truth for TC intensity, and 69 atmospheric variables are selected from the ERA5 data~\cite{era5}, with a 10$^{\circ}$ diameter area around the TC center extracted. The data from 1980 to 2017 is used for training, 2018 for validation, and 2019 to 2020 for testing. For real-time forecasting, we require FengWu forecast data. To address storage constraints, we use FengWu to generate 5-day forecasts for the ERA5 data in 2021 and 2022 and extract the corresponding TC-related variables. The PI is computed through future field calculations, including two-dimensional variables. All variables are normalized, as illustrated in:
\begin{eqnarray}
\begin{aligned}
\tilde{x} = \frac{(x-x_{mean} )}{x_{std}}.   
\end{aligned}
\end{eqnarray}
\subsubsection{Metrics.}
Consistent with most research, we use mean absolute error (MAE) as the performance evaluation metric, calculating the absolute error of $I$ and $\hat{I}$, with the formula:
\begin{eqnarray}
\begin{aligned}
MAE = \frac{1}{N}\sum_{i=1}^{N}\left |  I_{i}-{\hat{I}}_{i}  \right |.   
\end{aligned}
\end{eqnarray}
Meanwhile, the unit of MSW is commonly expressed in m$/$s, and we convert the experimental results according to the conversion of 1 knot $=$ 0.5144 m$/$s. Furthermore, when comparing the performance enhancement between different models, the forecast skill is often expressed as:
\begin{eqnarray}
sf(\%)=100\times \frac{eb-ef}{eb}, 
\end{eqnarray}
here, $eb$ represents the error of the baseline model, while $ef$ denotes the error of the model currently under evaluation. The value of $sf$ can be easily understood as the percentage reduction in forecast error compared to the baseline model.

\begin{table*}[]
\centering
\begin{tabular}{cllrrrrrlrrrrr}
\hline
\multirow{2}{*}{Basin}                                                         & \multicolumn{2}{c}{\multirow{2}{*}{Methods}} & \multicolumn{5}{c}{MSLP (hPa)}                                                                                                   &                      & \multicolumn{5}{c}{MSW (m$/$s)}                                                                                                    \\ \cline{4-8} \cline{10-14} 
                                                                               & \multicolumn{2}{c}{}                         & \multicolumn{1}{c}{24h} & \multicolumn{1}{c}{48h} & \multicolumn{1}{c}{72h} & \multicolumn{1}{c}{96h} & \multicolumn{1}{c}{120h} &                      & \multicolumn{1}{c}{24h} & \multicolumn{1}{c}{48h} & \multicolumn{1}{c}{72h} & \multicolumn{1}{c}{96h} & \multicolumn{1}{c}{120h} \\ \hline
\multirow{7}{*}{\begin{tabular}[c]{@{}c@{}}Global\\ 2019-\\ 2020\end{tabular}} & DL                     & TC-Pre              & 7.22                    & 11.78                   & 14.42                   & 15.71                   & 15.69                    &                      & 5.27                    & 8.52                    & 10.40                   & 11.11                   & 11.08                    \\
                                                                               &                        & MSCAR               & 7.13                    & 11.52                   & 14.33                   & 15.68                   & 15.37                    &                      & 5.56                    & 8.76                    & 10.72                   & 11.58                   & 11.31                    \\
                                                                               &                        & TCIF                & 7.94                    & 13.48                   & 16.31                   & 17.00                   & 15.97                    &                      & 6.28                    & 10.41                   & 12.38                   & 12.90                   & 11.89                    \\
                                                                               & NWP                    & ECMWF               & 7.65                    & 9.12                    & 9.80                    & 10.35                   & 11.07                    &                      & 7.48                    & 8.22                    & 8.32                    & 8.05                    & 8.15                     \\
                                                                               &                        & NCEP                & 6.98                    & 8.48                    & 9.39                    & 10.56                   & 11.29                    &                      & 5.95                    & 6.54                    & 7.02                    & 7.53                    & 8.33                     \\
                                                                               & Base                   & ERA5                & 9.59                    & 10.40                   & 10.48                   & 10.03                   & 8.96                     & \multicolumn{1}{r}{} & 17.72                   & 18.50                   & 18.52                   & 18.00                   & 16.86                    \\
                                                                               & Ours                   & VQLTI               & \textbf{5.87}           & \textbf{6.81}           & \textbf{7.09}           & \textbf{7.15}           & \textbf{6.64}            &                      & \textbf{4.30}           & \textbf{4.86}           & \textbf{5.06}           & \textbf{5.18}           & \textbf{5.05}            \\ \hline
\multirow{4}{*}{\begin{tabular}[c]{@{}c@{}}WP\\ 2019\end{tabular}}             & DL                     & MMSTN               & 9.68                    & 19.18                   & 26.64                   & 29.08                   & 28.62                    &                      & 5.73                    & 11.34                   & 16.23                   & 18.92                   & 19.94                    \\
                                                                               &                        & MGTCF               & 10.20                   & 15.07                   & 14.79                   & 13.32                   & 14.26                    &                      & 5.55                    & 8.25                    & 8.07                    & 8.01                    & 8.24                     \\
                                                                               & Official               & JTWC                & ---                     & ---                     & ---                     & ---                     & ---                      &                      & 4.60                    & 7.30                    & 9.70                    & 9.80                    & 9.10                     \\
                                                                               & Ours                   & VQLTI               & \textbf{6.62}           & \textbf{8.05}           & \textbf{8.96}           & \textbf{8.99}           & \textbf{8.09}            &                      & \textbf{4.24}           & \textbf{4.97}           & \textbf{5.43}           & \textbf{5.56}           & \textbf{5.23}            \\ \hline
\end{tabular}
\caption{Comparison of long-term TC intensity forecast error with MAE as the metric. The comparison is shown for the global and Western North Pacific Pacific (WP) basin. Note that MMSTN and MGTCF produce ensemble forecast results, and for a fair comparison, we calculate the average error, so the results may differ considerably from the original paper.}
\label{tab:long-term}
\end{table*}

\subsubsection{Implementation Details.}
VQLTI is trained using the PyTorch framework on an Nvidia A100 GPU. We employ the Adam optimization algorithm with a learning rate of 0.0001 and apply Exponential Moving Average (EMA) and L1 regularization with a coefficient of 0.00001. The batch size is set to 64, and the model is trained for 30 epochs.

In the first stage of training, the FPN extracts 256-dimensional feature maps with spatial scales of $\left \{ 40\times40, 20\times20, 10\times10, 10\times10  \right \} $ through convolutional networks~\cite{cnn}. These feature maps are then split into patches of size 5 and concatenated, followed by a linear layer to output a dimension of $d_{p}=128$. The codebook has a hidden space size $J=1024$, a dimension $d_{z}=128$, and the hyperparameter $\beta$ is 0.25. The cross-attention computation uses 3 heads and a dropout of 0.1.

In the second stage of training, the input consists of historical information from the past (-18h, -12h, -6h, 0h), denoted by $n=4$, and the forecast covers the next 2 days, denoted by $m=8$. Thanks to the design of the VQLTI model, we can easily modify the value of $m$ and the corresponding FengWu forecast data to achieve forecasting at any desired time step without additional training.

\subsection{Long-term Forecast}
As shown in Table~\ref{tab:long-term}, we compare our global TC forecasts from 2019 to 2020 with some recent deep learning-based methods, including TC-Pre~\cite{TC-Pred}, MSCAR~\cite{MSCAR}, and TCIF~\cite{TCIF}. We also compare with the well-known NWP models ECMWF-IFS ~\cite{TIGGE}and NCEP-GFS~\cite{TIGGE}. 
To ensure the model achieves effective forecasting, we also present the error of the ERA5 data itself. Note that ERA5 does not have any forecast results. The error here is calculated based on the input of the future ERA5. For MSLP, the minimum value of the $msl$ in ERA5 is selected, and for MSW, the maximum value of $\sqrt{(u10)^{2}+(v10)^{2}} $ is used as an approximate substitute, although they are not entirely consistent. 
Meanwhile, in the Western North Pacific (WP) basin where TC activities are most active, we also make comparisons with the deep learning-based MMSTN~\cite{mmstn} and MGTCF~\cite{MGTCF}, as well as the operational forecasts of Joint Typhoon Warning Center (JTCW)~\cite{JTWC}. The deep learning-based methods mentioned mostly have fixed forecast horizons, so they are all set to a fixed 20-step forecast during training.

In the global long-term forecast comparison, VQLTI demonstrates SOTA performance in 24h to 120h forecasts. The forecast skill of VQLTI not only comprehensively outperforms other deep learning-based methods, but its error accumulation rate is also significantly lower than other deep learning-based methods. TC-Pre, MSCAR, and TCIF have very fast error accumulation rates in short and medium-term forecasts, with their 48h MSLP forecast errors increasing by 63.16$\%$, 61.51$\%$, and 69.77$\%$ respectively compared to the 24h errors, while VQLTI's error only increases by 16.01$\%$. Relative to the NWP models ECWMF-IFS and NCEP-GFS, VQLTI reduces the 24h to 120h MSW forecast error by 35.65$\%$-42.51$\%$ and 25.69$\%$-39.38$\%$. The forecast errors of the VQLTI model are all lower than those of ERA5, which demonstrates the effectiveness of the model's forecasts.

In the WP basin, we only perform comparisons in 2019 due to the unique data used by MMSTN and MGTCF~\cite{mmstn,MGTCF}. MMSTN and MGTCF employ deep learning-based ensemble forecasting methods, with the mean of multiple forecasts used as the final result for a fair comparison. The 120-hour MSW forecast error of the VQLTI model is reduced by 71.73$\%$ and 36.53$\%$ compared to MMSTN and MGTCF, respectively. Furthermore, VQLTI's forecast error is 42.53$\%$ lower than that of JTWC.

\begin{table*}[]
\centering
\begin{tabular}{llrrrrrrrrrrr}
\hline
\multicolumn{2}{c}{\multirow{2}{*}{Methods}} & \multicolumn{5}{c}{MSLP (hPa)}                                                                                                            & \multicolumn{1}{l}{} & \multicolumn{5}{c}{MSW (m$/$s)}                                                                                                             \\ \cline{3-7} \cline{9-13} 
\multicolumn{2}{c}{}                         & \multicolumn{1}{c}{24h}   & \multicolumn{1}{c}{48h}   & \multicolumn{1}{c}{72h}   & \multicolumn{1}{c}{96h}   & \multicolumn{1}{c}{120h}  & \multicolumn{1}{l}{} & \multicolumn{1}{c}{24h}   & \multicolumn{1}{c}{48h}   & \multicolumn{1}{c}{72h}   & \multicolumn{1}{c}{96h}   & \multicolumn{1}{c}{120h}  \\ \hline
DL           & TC-Pre             & 7.51                      & 11.15                     & 12.88                     & 13.18                     & 13.03                     &                      & 5.12                      & 7.75                      & 8.88                      & 9.18                      & 8.89                      \\
             & MSCAR                         & 7.64                      & 11.22                     & 13.18                     & 13.99                     & 14.63                     &                      & 5.72                      & 8.22                      & 9.25                      & 9.95                      & 9.86                      \\
             & TCIF                          & 8.11                      & 12.98                     & 15.74                     & 16.83                     & 17.43                     &                      & 6.22                      & 9.97                      & 11.52                     & 12.33                     & 12.18                     \\
NWP          & ECMWF                         & 7.49                      & 8.28                      & 8.54                      & 9.05                      & 9.66                      &                      & 8.99                      & 9.19                      & 9.10                      & 8.63                      & 8.62                      \\
             & NCEP                          & 6.31                      & 7.63                      & 8.77                      & 9.29                      & 10.18                     &                      & 5.72                      & 5.87                      & 6.10                      & 6.07                      & 6.45                      \\
Base         & FengWu                        & \multicolumn{1}{l}{11.25} & \multicolumn{1}{l}{12.65} & \multicolumn{1}{l}{12.85} & \multicolumn{1}{l}{12.80} & \multicolumn{1}{l}{12.90} & \multicolumn{1}{l}{} & \multicolumn{1}{l}{20.04} & \multicolumn{1}{l}{20.51} & \multicolumn{1}{l}{19.93} & \multicolumn{1}{l}{19.28} & \multicolumn{1}{l}{18.70} \\
Ours         & VQLTI                     & \textbf{6.01}             & \textbf{6.74}             & \textbf{6.54}             & \textbf{6.22}             & \textbf{6.37}             &                      & 4.51                      & \textbf{4.80}             & \textbf{4.69}             & \textbf{4.56}             & \textbf{4.60}             \\
             & VQLTI (real-time)         & 6.42                      & 7.98                      & 8.29                      & 7.97                      & 7.86                      &                      & \textbf{4.47}             & 5.58                      & 5.78                      & 5.69                      & 5.53                      \\ \hline
\end{tabular}
\caption{Comparison of real-time TC intensity forecast error with MAE as the metric. The comparison is shown for the global region in 2022.}
\label{tab:real-time}
\end{table*}

\begin{table*}[t]
\centering
\begin{tabular}{cccrrrrrlrrrrr}
\hline
\multirow{2}{*}{VQ} & \multirow{2}{*}{FengWu} & \multirow{2}{*}{PI} & \multicolumn{5}{c}{MSLP (hPa)}                                                                                                   &                      & \multicolumn{5}{c}{MSW (m$/$s)}                                                                                                    \\ \cline{4-8} \cline{10-14} 
                    &                         &                     & \multicolumn{1}{c}{24h} & \multicolumn{1}{c}{48h} & \multicolumn{1}{c}{72h} & \multicolumn{1}{c}{96h} & \multicolumn{1}{c}{120h} &                      & \multicolumn{1}{c}{24h} & \multicolumn{1}{c}{48h} & \multicolumn{1}{c}{72h} & \multicolumn{1}{c}{96h} & \multicolumn{1}{c}{120h} \\ \hline
                    &                         &                     & 22.68                   & 25.17                   & 25.45                   & 25.34                   & 25.10                    & \multicolumn{1}{r}{} & 15.47                   & 17.99                   & 18.25                   & 18.17                   & 17.84                    \\
\checkmark                   &                         &                     & 20.22                   & 22.06                   & 21.96                   & 21.70                   & 21.45                    &                      & 15.76                   & 17.10                   & 17.03                   & 16.77                   & 16.35                    \\
\checkmark                   & \checkmark                       &                     & 6.60                    & 8.56                    & 9.44                    & 10.24                   & 11.35                    & \multicolumn{1}{r}{} & 4.64                    & 5.92                    & 6.34                    & 6.96                    & 7.31                     \\
\checkmark                   & \checkmark                       & \checkmark                   & \textbf{6.42}           & \textbf{7.98}           & \textbf{8.29}           & \textbf{7.97}           & \textbf{7.86}            & \multicolumn{1}{r}{} & \textbf{4.47}           & \textbf{5.58}           & \textbf{5.78}           & \textbf{5.69}           & \textbf{5.53}            \\ \hline
\end{tabular}

\caption{Ablation experiments based on real-time forecasts in 2022 with MAE as the metric. VQ represents the discretization of latent variables by learning the codebook during training. FengWu represents the introduction of FengWu forecast field information. PI represents the calculation of PI data.}
\label{tab:ablation}
\end{table*}

\subsection{Real-time Forecast}
Real-time forecasting needs to address two key issues. First, the ERA5 data requires data assimilation and correction, and cannot be obtained in real-time, so we use relatively available real-time analysis data to replace the historical ERA5 inputs. Second, future ERA5 fields are also unavailable, so we substitute them with FengWu's forecast results. We use 2021 data to fine-tune VQLTI and evaluate it in 2022, as presented in Table~\ref{tab:real-time}.

VQLTI (real-time) continues to exhibit excellent performance in real-time forecasting. Its 24h forecast shows negligible changes compared to VQLTI and even experiences a slight improvement in 24h MSW forecasting. VQLTI (real-time)'s long-term forecasting does exhibit some performance degradation, with its 120h MSW forecast error increasing by 20.22$\%$ compared to VQLTI. We consider this error reasonable, as FengWu's forecasts have inherent discrepancies compared to ERA5 (as shown in Figure ~\ref{era5_fengwu_diff}), and the errors in FengWu's long-term forecasts lead to deviations in the TC positions obtained by the ECMWF's TC tracker, resulting in misalignment of the extracted spatial information. Although VQLTI (real-time) long-term forecasting has declined to some extent, its forecasting skill remains superior to other deep learning-based methods. Compared to TC-Pre, MSCAR, and TCIF, its 120h MSW forecast error decreases by 37.80$\%$, 43.91$\%$, and 54.60$\%$ respectively. Compared to the NWP models ECMWF-IFS and NCEP-GFS, VQLTI (real-time)'s 120h MSW forecast error decreases by 35.85$\%$ and 14.26$\%$. We have also calculated the error of the FengWu forecast data, and the VQLTI's forecast skill is significantly better than FengWu, which verifies the effectiveness of the fine-tuning.

\begin{figure}[t]
\centering
\includegraphics[width=0.45\textwidth]{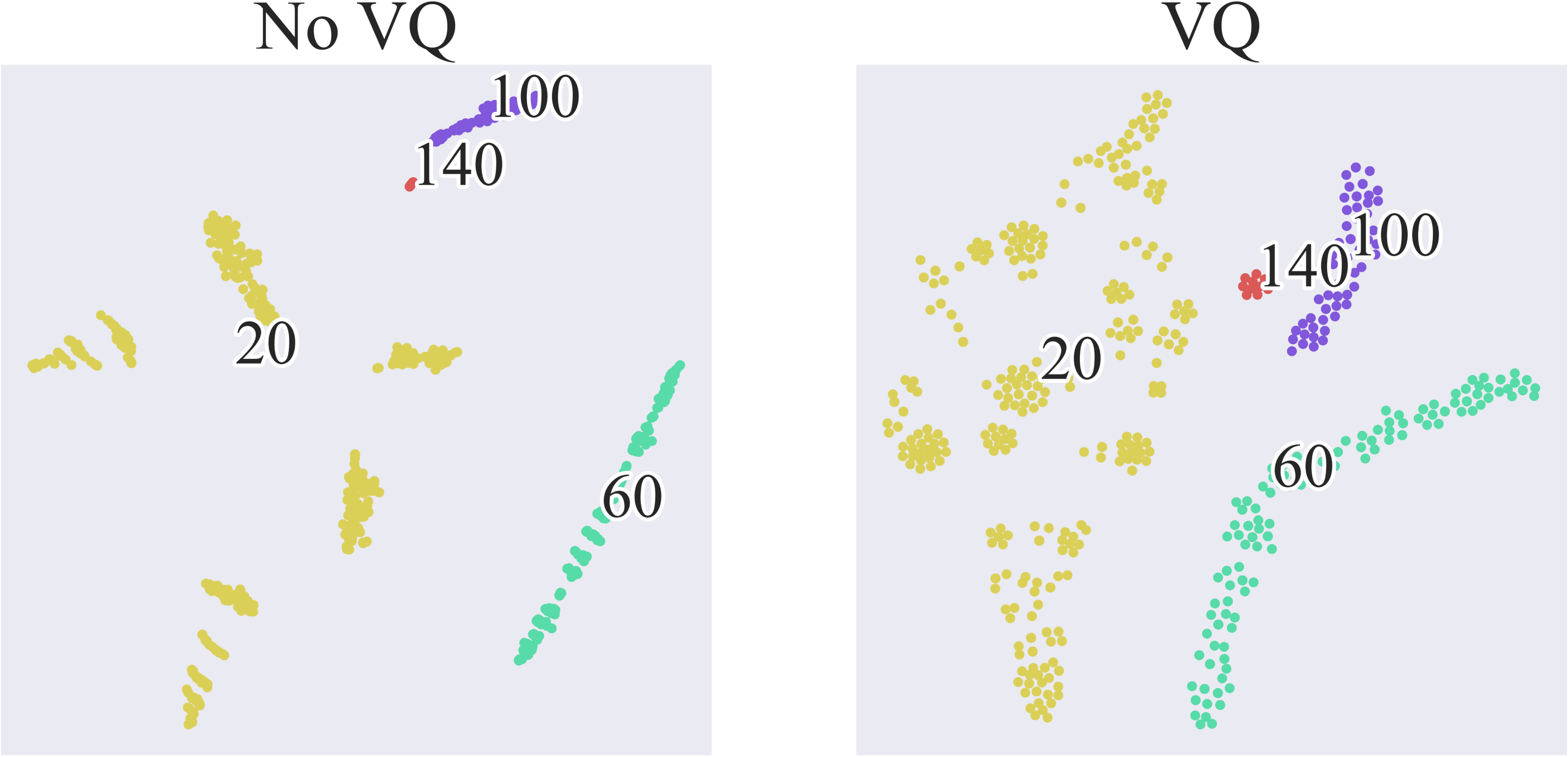} 
\caption{T-SNE visualization of latent variables. The same color represents the same intensity value, and here we use MSW (knot) as an example.}
\label{VQ_rule}
\end{figure}

\subsection{Ablation Studies}
This ablation study explores the design rationale of the VQLTI model. As shown in Table ~\ref{tab:ablation}, it explores whether to use a codebook to quantify latent variables (VQ), whether to introduce FengWu forecast data (FengWu), and whether to introduce PI data (PI). The experimental results show that using codebook for quantization of latent variables is beneficial for long-term forecasting, reducing the MSLP forecasting error by 14.54$\%$ for the 120h forecast. FengWu's forecast field plays a huge role, and introducing the FengWu forecast field, the error in the 24h to 120h MSW forecast is reduced by 55.29$\%$-70.56$\%$. The constraint brought by PI also plays a significant role in improving the long-term forecast skill, reducing the MSLP and MSW forecast errors by 30.75$\%$ and 24.35$\%$ for the 120h forecast, respectively.

Furthermore, we visualize the latent variables $Z$ of VQLTI with and without codebook quantization during training using t-SNE~\cite{tsne}, as shown in Figure~\ref{VQ_rule}. We use the same color to represent the same intensity, and we find that the latent variables quantized using codebook not only correctly represent the same intensity, but also maintain a certain intra-class distance. This is consistent with our original intention, where the codebook successfully transfers the TC intensity information to the discrete latent space while ensuring the difference caused by the difference in spatial information between the same intensities.

In summary, these results validate the rationality and effectiveness of the VQLTI model design. For additional validation experiments and further details, please refer to the Appendix.

\section{Conclusion}
This study presents VQLTI, a model that achieves highly accurate real-time global long-term tropical cyclone (TC) intensity forecasting. We transfer the TC intensity information into a discrete latent space and fully leverage the forecast fields of the machine learning-based weather prediction model FengWu as spatial information conditions to compensate for the lack of physical knowledge learning in the VQLTI model. Finally, we propose to compute the potential intensity (PI) to provide physical constraints, mitigating the underestimation of TC intensity by the forecast fields.
To the best of our knowledge, this is the first real-time TC intensity forecasting model that utilizes the forecast fields of a machine learning-based weather prediction model, and it performs the forecasting in the discrete latent space, resulting in low inference cost and no need for post-processing by forecasters. 
The experimental results demonstrate the superior performance and the rationality of the design of VQLTI.

\section{Acknowledgments}
This research is supported by the National Key R\&D Program of China (Grant No.2024YFB4613400), the Anhui Provincial Natural Science Foundation (Grant No.2408085QF214), the Fundamental Research Funds for the Central Universities (Grant No. WK2100000045), the Opening Project of the State Key Laboratory of General Artificial Intelligence(Grant No. SKLAGI2024OP10, Grant No.SKLAGI2024OP11), the National Natural Science Foundation of China under grand No.U19B2044 and the Shanghai Science and Technology Commission Project (23DZ1204704). 

\bibliography{aaai25}

\begin{thebibliography}{45}
\providecommand{\natexlab}[1]{#1}

\bibitem[{Bao et~al.(2017)Bao, Chen, Wen, Li, and Hua}]{CVAE-GAN}
Bao, J.; Chen, D.; Wen, F.; Li, H.; and Hua, G. 2017.
\newblock CVAE-GAN: fine-grained image generation through asymmetric training.
\newblock In \emph{Proceedings of the IEEE international conference on computer vision}, 2745--2754.

\bibitem[{Bi et~al.(2023)Bi, Xie, Zhang, Chen, Gu, and Tian}]{pangu}
Bi, K.; Xie, L.; Zhang, H.; Chen, X.; Gu, X.; and Tian, Q. 2023.
\newblock Accurate medium-range global weather forecasting with 3D neural networks.
\newblock \emph{Nature}, 619(7970): 533--538.

\bibitem[{Bougeault et~al.(2010)Bougeault, Toth, Bishop, Brown, Burridge, Chen, Ebert, Fuentes, Hamill, Mylne et~al.}]{TIGGE}
Bougeault, P.; Toth, Z.; Bishop, C.; Brown, B.; Burridge, D.; Chen, D.~H.; Ebert, B.; Fuentes, M.; Hamill, T.~M.; Mylne, K.; et~al. 2010.
\newblock The THORPEX interactive grand global ensemble.
\newblock \emph{Bulletin of the American Meteorological Society}, 91(8): 1059--1072.

\bibitem[{Bourdin et~al.(2022)Bourdin, Fromang, Dulac, Cattiaux, and Chauvin}]{low1}
Bourdin, S.; Fromang, S.; Dulac, W.; Cattiaux, J.; and Chauvin, F. 2022.
\newblock Intercomparison of four tropical cyclones detection algorithms on ERA5.

\bibitem[{Buizza et~al.(2018)Buizza, Balmaseda, Brown, English, Forbes, Geer, Haiden, Leutbecher, Magnusson, Rodwell et~al.}]{nwp}
Buizza, R.; Balmaseda, M.~A.; Brown, A.; English, S.; Forbes, R.; Geer, A.; Haiden, T.; Leutbecher, M.; Magnusson, L.; Rodwell, M.; et~al. 2018.
\newblock \emph{The development and evaluation process followed at ECMWF to upgrade the Integrated Forecasting System (IFS)}.
\newblock European Centre for Medium Range Weather Forecasts.

\bibitem[{Chen et~al.(2019)Chen, Zhang, Bai, and Wan}]{JTWC}
Chen, G.; Zhang, X.; Bai, L.; and Wan, R. 2019.
\newblock Verification of tropical cyclone operational forecast in 2018.
\newblock \emph{Proceedings of the ESCAP/WMO Typhoon Committee, Guangzhou, China}, 26.

\bibitem[{Chen et~al.(2023{\natexlab{a}})Chen, Han, Gong, Bai, Ling, Luo, Chen, Ma, Zhang, Su et~al.}]{fengwu}
Chen, K.; Han, T.; Gong, J.; Bai, L.; Ling, F.; Luo, J.-J.; Chen, X.; Ma, L.; Zhang, T.; Su, R.; et~al. 2023{\natexlab{a}}.
\newblock Fengwu: Pushing the skillful global medium-range weather forecast beyond 10 days lead.
\newblock \emph{arXiv preprint arXiv:2304.02948}.

\bibitem[{Chen et~al.(2023{\natexlab{b}})Chen, Zhong, Zhang, Cheng, Xu, Qi, and Li}]{fuxi}
Chen, L.; Zhong, X.; Zhang, F.; Cheng, Y.; Xu, Y.; Qi, Y.; and Li, H. 2023{\natexlab{b}}.
\newblock FuXi: A cascade machine learning forecasting system for 15-day global weather forecast.
\newblock \emph{npj Climate and Atmospheric Science}, 6(1): 190.

\bibitem[{Dosovitskiy et~al.(2020)Dosovitskiy, Beyer, Kolesnikov, Weissenborn, Zhai, Unterthiner, Dehghani, Minderer, Heigold, Gelly et~al.}]{ViT}
Dosovitskiy, A.; Beyer, L.; Kolesnikov, A.; Weissenborn, D.; Zhai, X.; Unterthiner, T.; Dehghani, M.; Minderer, M.; Heigold, G.; Gelly, S.; et~al. 2020.
\newblock An image is worth 16x16 words: Transformers for image recognition at scale.
\newblock \emph{arXiv preprint arXiv:2010.11929}.

\bibitem[{Emanuel(2003)}]{define}
Emanuel, K. 2003.
\newblock Tropical cyclones.
\newblock \emph{Annual review of earth and planetary sciences}, 31(1): 75--104.

\bibitem[{Emanuel(2018)}]{PI_100year}
Emanuel, K. 2018.
\newblock 100 years of progress in tropical cyclone research.
\newblock \emph{Meteorological Monographs}, 59(1): 15--1.

\bibitem[{Emanuel(1986)}]{PI_ori}
Emanuel, K.~A. 1986.
\newblock An air-sea interaction theory for tropical cyclones. Part I: Steady-state maintenance.
\newblock \emph{Journal of Atmospheric Sciences}, 43(6): 585--605.

\bibitem[{Emanuel(1987)}]{PI_name}
Emanuel, K.~A. 1987.
\newblock The dependence of hurricane intensity on climate.
\newblock \emph{Nature}, 326(6112): 483--485.

\bibitem[{Gilford(2021)}]{PyPI}
Gilford, D.~M. 2021.
\newblock pyPI (v1. 3): Tropical cyclone potential intensity calculations in Python.
\newblock \emph{Geoscientific Model Development}, 14(5): 2351--2369.

\bibitem[{Graves and Graves(2012)}]{LSTM}
Graves, A.; and Graves, A. 2012.
\newblock Long short-term memory.
\newblock \emph{Supervised sequence labelling with recurrent neural networks}, 37--45.

\bibitem[{Haiden et~al.(2018)Haiden, Janousek, Vitart, Bouall{\`e}gue, Ferranti, Prates, and Richardson}]{ECMWF-IFS}
Haiden, T.; Janousek, M.; Vitart, F.; Bouall{\`e}gue, Z.~B.; Ferranti, L.; Prates, F.; and Richardson, D. 2018.
\newblock \emph{Evaluation of ECMWF forecasts, including the 2018 upgrade}.
\newblock European Centre for Medium Range Weather Forecasts Reading, UK.

\bibitem[{Han et~al.(2024)Han, Guo, Ling, Chen, Gong, Luo, Gu, Dai, Ouyang, and Bai}]{fengwu_ghr}
Han, T.; Guo, S.; Ling, F.; Chen, K.; Gong, J.; Luo, J.; Gu, J.; Dai, K.; Ouyang, W.; and Bai, L. 2024.
\newblock Fengwu-ghr: Learning the kilometer-scale medium-range global weather forecasting.
\newblock \emph{arXiv preprint arXiv:2402.00059}.

\bibitem[{Hersbach et~al.(2020)Hersbach, Bell, Berrisford, Hirahara, Hor{\'a}nyi, Mu{\~n}oz-Sabater, Nicolas, Peubey, Radu, Schepers et~al.}]{era5}
Hersbach, H.; Bell, B.; Berrisford, P.; Hirahara, S.; Hor{\'a}nyi, A.; Mu{\~n}oz-Sabater, J.; Nicolas, J.; Peubey, C.; Radu, R.; Schepers, D.; et~al. 2020.
\newblock The ERA5 global reanalysis.
\newblock \emph{Quarterly Journal of the Royal Meteorological Society}, 146(730): 1999--2049.

\bibitem[{Hsu et~al.(2024)Hsu, Liu, Peng, Chen, Chang, Hsiao, Fong, Hong, Cheng, Lu et~al.}]{five—wfmodel-error}
Hsu, K.; Liu, C.-C.; Peng, M.; Chen, D.-S.; Chang, P.-L.; Hsiao, L.-F.; Fong, C.-T.; Hong, J.-S.; Cheng, C.-P.; Lu, K.-C.; et~al. 2024.
\newblock Performance of Five Machine Learning-based Global Weather Prediction Models in the East Asia Region.

\bibitem[{Huang et~al.(2022)Huang, Bai, Chan, and Zhang}]{mmstn}
Huang, C.; Bai, C.; Chan, S.; and Zhang, J. 2022.
\newblock MMSTN: A Multi-Modal Spatial-Temporal Network for Tropical Cyclone Short-Term Prediction.
\newblock \emph{Geophysical Research Letters}, 49(4): e2021GL096898.

\bibitem[{Huang et~al.(2023)Huang, Bai, Chan, Zhang, and Wu}]{MGTCF}
Huang, C.; Bai, C.; Chan, S.; Zhang, J.; and Wu, Y. 2023.
\newblock MGTCF: multi-generator tropical cyclone forecasting with heterogeneous meteorological data.
\newblock In \emph{Proceedings of the AAAI Conference on Artificial Intelligence}, volume~37, 5096--5104.

\bibitem[{Jiang et~al.(2023)Jiang, Zhang, Hu, Wu, Liu, Xiao, and Duan}]{transformer-based}
Jiang, W.; Zhang, D.; Hu, G.; Wu, T.; Liu, L.; Xiao, Y.; and Duan, Z. 2023.
\newblock Transformer-based tropical cyclone track and intensity forecasting.
\newblock \emph{Journal of Wind Engineering and Industrial Aerodynamics}, 238: 105440.

\bibitem[{Knapp et~al.(2010)Knapp, Kruk, Levinson, Diamond, and Neumann}]{IBTrACS}
Knapp, K.~R.; Kruk, M.~C.; Levinson, D.~H.; Diamond, H.~J.; and Neumann, C.~J. 2010.
\newblock The international best track archive for climate stewardship (IBTrACS) unifying tropical cyclone data.
\newblock \emph{Bulletin of the American Meteorological Society}, 91(3): 363--376.

\bibitem[{Krizhevsky, Sutskever, and Hinton(2012)}]{cnn}
Krizhevsky, A.; Sutskever, I.; and Hinton, G.~E. 2012.
\newblock Imagenet classification with deep convolutional neural networks.
\newblock \emph{Advances in neural information processing systems}, 25.

\bibitem[{Lam et~al.(2023)Lam, Sanchez-Gonzalez, Willson, Wirnsberger, Fortunato, Alet, Ravuri, Ewalds, Eaton-Rosen, Hu et~al.}]{graph_cast}
Lam, R.; Sanchez-Gonzalez, A.; Willson, M.; Wirnsberger, P.; Fortunato, M.; Alet, F.; Ravuri, S.; Ewalds, T.; Eaton-Rosen, Z.; Hu, W.; et~al. 2023.
\newblock Learning skillful medium-range global weather forecasting.
\newblock \emph{Science}, 382(6677): 1416--1421.

\bibitem[{Lin et~al.(2017)Lin, Doll{\'a}r, Girshick, He, Hariharan, and Belongie}]{FPN}
Lin, T.-Y.; Doll{\'a}r, P.; Girshick, R.; He, K.; Hariharan, B.; and Belongie, S. 2017.
\newblock Feature pyramid networks for object detection.
\newblock In \emph{Proceedings of the IEEE conference on computer vision and pattern recognition}, 2117--2125.

\bibitem[{Magnusson et~al.(2021)Magnusson, Majumdar, Emerton, Richardson, Alonso-Balmaseda, Baugh, Bechtold, Bidlot, Bonanni, Bonavita, Bormann, Brown, Browne, Carr, Dahoui, Chiara, Diamantakis, Duncan, English, Forbes, Geer, Haiden, Healy, Hewson, Ingleby, Janousek, Kuehnlein, Lang, Lock, McNally, Mogensen, Pappenberger, Polichtchouk, Prates, Prudhomme, Rabier, de~Rosnay, Quintino, and Rennie}]{tracker}
Magnusson, L.; Majumdar, S.; Emerton, R.; Richardson, D.; Alonso-Balmaseda, M.; Baugh, C.; Bechtold, P.; Bidlot, J.-R.; Bonanni, A.; Bonavita, M.; Bormann, N.; Brown, A.; Browne, P.; Carr, H.; Dahoui, M.; Chiara, G.~D.; Diamantakis, M.; Duncan, D.; English, S.; Forbes, R.; Geer, A.; Haiden, T.; Healy, S.; Hewson, T.; Ingleby, B.; Janousek, M.; Kuehnlein, C.; Lang, S.; Lock, S.-J.; McNally, T.; Mogensen, K.; Pappenberger, F.; Polichtchouk, I.; Prates, F.; Prudhomme, C.; Rabier, F.; de~Rosnay, P.; Quintino, T.; and Rennie, M. 2021.
\newblock Tropical cyclone activities at ECMWF.

\bibitem[{Malakar et~al.(2020)Malakar, Kesarkar, Bhate, Singh, and Deshamukhya}]{low2}
Malakar, P.; Kesarkar, A.; Bhate, J.; Singh, V.; and Deshamukhya, A. 2020.
\newblock Comparison of reanalysis data sets to comprehend the evolution of tropical cyclones over North Indian Ocean.
\newblock \emph{Earth and Space Science}, 7(2): e2019EA000978.

\bibitem[{Meng et~al.(2023)Meng, Yang, Yao, Wang, and Song}]{PTCIF}
Meng, F.; Yang, K.; Yao, Y.; Wang, Z.; and Song, T. 2023.
\newblock Tropical cyclone intensity probabilistic forecasting system based on deep learning.
\newblock \emph{International Journal of Intelligent Systems}, 2023(1): 3569538.

\bibitem[{Pan, Xu, and Shi(2019)}]{RNN-base}
Pan, B.; Xu, X.; and Shi, Z. 2019.
\newblock Tropical cyclone intensity prediction based on recurrent neural networks.
\newblock \emph{Electronics Letters}, 55(7): 413--415.

\bibitem[{Rasul et~al.(2024)Rasul, Bennett, Vicente, Gupta, Ghonia, Schneider, and Nevmyvaka}]{vq-tr}
Rasul, K.; Bennett, A.; Vicente, P.; Gupta, U.; Ghonia, H.; Schneider, A.; and Nevmyvaka, Y. 2024.
\newblock {VQ}-{TR}: Vector Quantized Attention for Time Series Forecasting.
\newblock In \emph{The Twelfth International Conference on Learning Representations}.

\bibitem[{Rasul et~al.(2022)Rasul, Park, Ramstr{\"o}m, and Kim}]{vq-ar}
Rasul, K.; Park, Y.-J.; Ramstr{\"o}m, M.~N.; and Kim, K.-M. 2022.
\newblock Vq-ar: Vector quantized autoregressive probabilistic time series forecasting.
\newblock \emph{arXiv preprint arXiv:2205.15894}.

\bibitem[{R{\"u}ttgers et~al.(2019)R{\"u}ttgers, Lee, Jeon, and You}]{cost}
R{\"u}ttgers, M.; Lee, S.; Jeon, S.; and You, D. 2019.
\newblock Prediction of a typhoon track using a generative adversarial network and satellite images.
\newblock \emph{Scientific reports}, 9(1): 6057.

\bibitem[{Shen et~al.(2023)Shen, Su, Zhang, and Hu}]{cma-gfs}
Shen, X.; Su, Y.; Zhang, H.; and Hu, J. 2023.
\newblock New version of the CMA-GFS dynamical core based on the predictor--corrector time integration scheme.
\newblock \emph{Journal of Meteorological Research}, 37(3): 273--285.

\bibitem[{Sohn, Lee, and Yan(2015)}]{CVAE}
Sohn, K.; Lee, H.; and Yan, X. 2015.
\newblock Learning structured output representation using deep conditional generative models.
\newblock \emph{Advances in neural information processing systems}, 28.

\bibitem[{Van Den~Oord, Vinyals et~al.(2017)}]{VQVAE}
Van Den~Oord, A.; Vinyals, O.; et~al. 2017.
\newblock Neural discrete representation learning.
\newblock \emph{Advances in neural information processing systems}, 30.

\bibitem[{Van~der Maaten and Hinton(2008)}]{tsne}
Van~der Maaten, L.; and Hinton, G. 2008.
\newblock Visualizing data using t-SNE.
\newblock \emph{Journal of machine learning research}, 9(11).

\bibitem[{Vaswani et~al.(2017)Vaswani, Shazeer, Parmar, Uszkoreit, Jones, Gomez, Kaiser, and Polosukhin}]{transformer}
Vaswani, A.; Shazeer, N.; Parmar, N.; Uszkoreit, J.; Jones, L.; Gomez, A.~N.; Kaiser, {\L}.; and Polosukhin, I. 2017.
\newblock Attention is all you need.
\newblock \emph{Advances in neural information processing systems}, 30.

\bibitem[{Wang, Li, and Zheng(2024)}]{TCIF}
Wang, C.; Li, X.; and Zheng, G. 2024.
\newblock Tropical cyclone intensity forecasting using model knowledge guided deep learning model.
\newblock \emph{Environmental Research Letters}, 19(2): 024006.

\bibitem[{Wang et~al.(2024)Wang, Chen, Liu, Han, Li, and Bai}]{MSCAR}
Wang, X.; Chen, K.; Liu, L.; Han, T.; Li, B.; and Bai, L. 2024.
\newblock Global Tropical Cyclone Intensity Forecasting with Multi-modal Multi-scale Causal Autoregressive Model.
\newblock \emph{arXiv preprint arXiv:2402.13270}.

\bibitem[{Woodruff, Irish, and Camargo(2013)}]{flooding}
Woodruff, J.~D.; Irish, J.~L.; and Camargo, S.~J. 2013.
\newblock Coastal flooding by tropical cyclones and sea-level rise.
\newblock \emph{Nature}, 504(7478): 44--52.

\bibitem[{Yang et~al.(2019)Yang, Li, Zhao, Wang, Wang, Sou, Yang, Hu, Tang, Mok et~al.}]{Mangkhut}
Yang, J.; Li, L.; Zhao, K.; Wang, P.; Wang, D.; Sou, I.~M.; Yang, Z.; Hu, J.; Tang, X.; Mok, K.~M.; et~al. 2019.
\newblock A comparative study of Typhoon Hato (2017) and Typhoon Mangkhut (2018)—Their impacts on coastal inundation in Macau.
\newblock \emph{Journal of Geophysical Research: Oceans}, 124(12): 9590--9619.

\bibitem[{Zhang, Fang, and Yu(2023)}]{zhang2023forecast}
Zhang, X.; Fang, J.; and Yu, Z. 2023.
\newblock The forecast skill of tropical cyclone genesis in two global ensembles.
\newblock \emph{Weather and Forecasting}, 38(1): 83--97.

\bibitem[{Zhang et~al.(2022)Zhang, Yang, Shi, Wang, Du, Zhang, and Liu}]{TC-Pred}
Zhang, Z.; Yang, X.; Shi, L.; Wang, B.; Du, Z.; Zhang, F.; and Liu, R. 2022.
\newblock A neural network framework for fine-grained tropical cyclone intensity prediction.
\newblock \emph{Knowledge-Based Systems}, 241: 108195.

\bibitem[{Zhou and Juang(2023)}]{NCEP-GFS}
Zhou, X.; and Juang, H.-M.~H. 2023.
\newblock A Model Instability Issue in the NCEP Global Forecast System Version 16 and Potential Solutions.
\newblock \emph{EGUsphere}, 2023: 1--22.

\end{thebibliography}

\end{document}